\documentclass{article}
\usepackage{caption}
\usepackage{amssymb}
\usepackage{amsmath}
\usepackage{xcolor}
\usepackage{algorithm}
\usepackage{algpseudocode}
\usepackage{wrapfig}
\usepackage{mathptmx}
\usepackage{graphicx}
\usepackage{booktabs}
\usepackage{paralist}
\usepackage{pifont}   %
\usepackage{multirow} %

\definecolor{RubineRed}{RGB}{209,0,86}

\definecolor{R1Color}{RGB}{209,0,86}
\definecolor{R2Color}{RGB}{37,150,190}
\definecolor{R3Color}{RGB}{38,156,142}
\definecolor{R4Color}{RGB}{145,104,72}

\newcommand{\cai}[1]{\textcolor{R1Color}{\textbf{cai3}}}
\newcommand{\RZcW}[1]{\textcolor{R2Color}{\textbf{RZcW}}}
\newcommand{\bdgn}[1]{\textcolor{R3Color}{\textbf{bdgn}}}
\newcommand{\Kpx}[1]{\textcolor{R4Color}{\textbf{Kpx6}}}

\definecolor{HopBine}{RGB}{170,157,46}
\definecolor{RougeWave}{RGB}{13,82,87}
\definecolor{SolarFlare}{RGB}{211,131,43}
\newcommand{\ourcolor}[1]{\textcolor{SolarFlare}{#1}}

\newcommand{\ours}{\ourcolor{\texttt{PEGrad}}}

\newcommand{\PM}[1]{\scriptsize{~$\pm$#1}}

\usepackage[preprint]{corl_2025} %

\title{Non-conflicting Energy Minimization \\ in Reinforcement Learning based Robot Control}

\author{
  Skand Peri\hspace{1em} Akhil Perincherry\footnotemark[1]\hspace{1em} Bikram Pandit\footnotemark[1]\hspace{1em} Stefan Lee\\ 
  Oregon State University \\ 
  \textbf{Project Page: \href{https://pvskand.github.io/projects/PEGRAD}{https://pvskand.github.io/projects/PEGrad}} 
}

\begin{document}
\maketitle
{\renewcommand{\thefootnote}{}\footnotetext{\scalebox{0.95}{\noindent\hspace{-1em} $^*$Equal contribution.}}}

\begin{abstract}
    Efficient robot control often requires balancing task performance with energy expenditure. A common approach in reinforcement learning (RL) is to penalize energy use directly as part of the reward function. This requires carefully tuning weight terms to avoid undesirable trade-offs where energy minimization harms task success. In this work, we propose a hyperparameter-free gradient optimization method to minimize energy expenditure without conflicting with task performance. Inspired by recent works in multitask learning, our method applies policy gradient projection between task and energy objectives to derive policy updates that minimize energy expenditure in ways that do not impact task performance. We evaluate this technique on standard locomotion benchmarks of DM-Control and HumanoidBench and demonstrate a reduction of $64\%$ energy usage while maintaining comparable task performance. Further, we conduct experiments on a Unitree GO2 quadruped showcasing Sim2Real transfer of energy efficient policies. Our method is easy to implement in standard RL pipelines with minimal code changes, is applicable to any policy gradient method, and offers a principled alternative to reward shaping for energy efficient control policies.
\end{abstract}


\section{Introduction}
\vspace{-5pt}

Untethered robots are inherently constrained by their battery capacity limiting their deployment duration. For example, the Unitree Go2 \citep{unitree-go2} typically operates for only 1–4 hours per charge under low-speed locomotion using its factory controllers, with charging times ranging from 1–2 hours. To maximize operational time, control policies must minimize energy expenditure while still ensuring task completion -- any excess energy use shortens battery life while insufficient effort may result in higher rates of task failure. Furthermore, high-energy behaviors can pose safety risks to the robot and its environment, such as excessive actuator wear or destabilizing interactions with the ground.

\begin{figure}[t]
\hspace{0.1in}\includegraphics[width=0.925\linewidth]{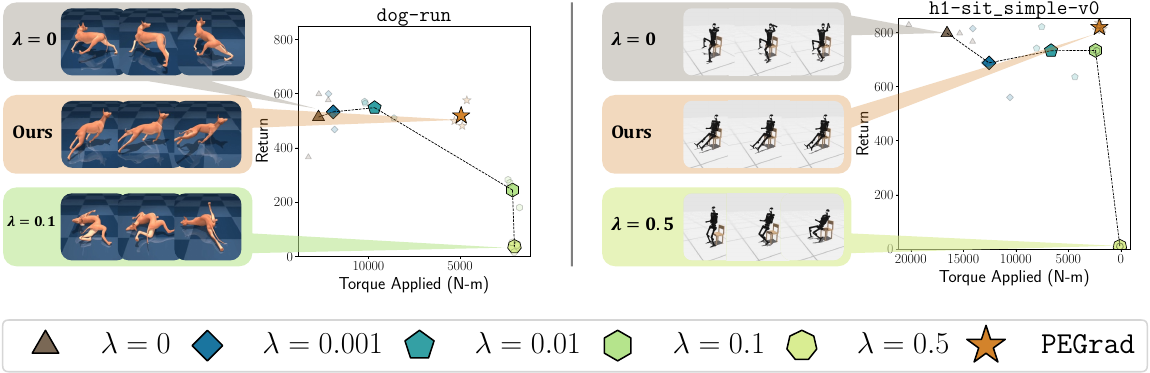}
  \caption{RL control policies often optimize a weighted combination of task reward and energy penalties, i.e., $r_{\text{task}} + \lambda r_{\text{energy}}$. However, tuning the weighting factor $\lambda$ is challenging due to high variability in its optimal value across tasks, environments, and embodiments. \textit{(Left)} When a Soft Actor Critic (SAC) agent is trained on the \texttt{dog-run} task,  $\lambda{=}0.01$ and $\lambda{=}0.1$ result in significantly different performance -- with the policy at $\lambda{=}0.1$ achieving low returns by crawling rather than running. However, $\lambda{=}0.1$ works well in the less dynamic \texttt{dog-walk} environment \textit{(Not shown)}. \textit{(Right)} For a humanoid \texttt{sitting} task, both $\lambda{=}0.01$ and $0.1$ yield policies that are equally energy-efficient and task-effective, showcasing the inter-environment variability. In both cases, our proposed hyperparameter-free method, \ours{} (\textcolor[HTML]{D4853B}{$\bigstar$}), leads to  performant and energy efficient policies. (The lightly shaded markers represent the checkpoint with the highest average evaluation score over 50 episodes, while the solid markers indicate the overall average value across all runs.) }
  \label{fig:00_intro}
  \vspace{-11pt}
\end{figure}

One of the most common and increasingly popular ways of learning robot control policies has been using reinforcement learning (RL) \cite{hwangbo2019learning, lee2020learning, long2023robust, yang2022fast, kumar2021rma, siekmann2020sim, van2024revisiting} -- especially for complex and highly dynamic embodiments like legged robots. During policy training, energy minimization is typically incorporated into the reward function through penalties on joint torques \cite{kumar2021rma, van2024revisiting} or mechanical work \cite{pmlr-v205-fu23a, tan2018sim}. A common formulation is a weighted sum: $r = r_{\text{task}} + \lambda r_{\text{energy}}$, where $r_{\text{task}}$ is the task specific reward and $r_{\text{energy}}$ is the energy penalty. The $\lambda$ coefficient then becomes a critical hyperparameter for balancing between these (often competing) objectives. However, different embodiments and tasks naturally demand different effort, e.g., running being more energetic than standing. Consequentially, the performance of learned policies can be highly sensitive to the choice of $\lambda$, making it difficult to tune in practice. As shown in the left plot in Figure \ref{fig:00_intro}, different values of $\lambda$ offer different trade-offs between a trained policy's average return and overall motor torque for a quadruped running task. While $\lambda$'s of 0, 0.001, and 0.01 all produce similar returns (though differing torques), higher values begin to significantly degrade task performance as energy minimization starts to dominate.

Rather than tuning a trade-off between energy expenditure and task performance, we would ideally like to specify an ordering of these objectives. That is to say, we would like robots to expend the least amount of energy \textbf{\textit{necessary}} to successfully complete the task --  adapting energy use to task demands without requiring per-task manual tuning. To address this, we propose \ours{} -- a hyperparameter-free method that trains policies that simultaneously achieve high task performance and low energy expenditure. As shown by the orange stars (\textcolor{SolarFlare}{$\bigstar$}) Figure \ref{fig:00_intro}, \ours{}-based policies can automatically find performant policies with low energy expenditure across different tasks and embodiments. 

 Our key idea is to formulate policy training as a multi-objective optimization problem and derive a descent direction for policy parameters which minimize energy by moving along approximate level-sets of the reward function. Specifically, we modify the gradients of energy with respect to the policy parameters to be orthogonal to the task reward gradient at each time step of optimization. We conduct experiments on various locomotion benchmarks such as \texttt{DM-Control} suite, \texttt{HumanoidBench}, and \texttt{IssacLab} that show the efficacy of \ours{}. Specifically, we show an average of $64\%$ energy-efficiency on these benchmarks compared to vanilla RL policy ($\lambda=0$). Finally, we transfer a learned policy to a real robot to demonstrate real-world improvements to battery life from running our policies.

\textbf{Contributions.} We summarize our main contributions:\\[2pt]
- We propose \ours{}, a hyperparameter-free gradient optimization method that minimizes energy expenditure without conflicting with task performance in policy gradient RL methods.\\[2pt]
- We experiment in two simulated benchmarks, DM-Control \cite{tunyasuvunakool2020dm_control} and HumanoidBench \cite{sferrazza2024humanoidbench}, and show that \ours{} decreases energy expenditure by $64\%$ while maintaining task performance.\\[2pt]
- We demonstrate the Sim2Real transfer of \ours{} on Unitree GO2 robot and evaluate energy efficiency in terms of the current drawn from the battery and the net torque applied by the robot while executing Standing and Walking (SaW) policies.

\section{Related Works}
\vspace{-5pt}
\textbf{Multi-objective optimization}: Multi-objective optimization (MOO) studies the problem of optimizing a set of potentially conflicting objectives. We focus on MOO methods that explicitly manipulate gradients to mitigate negative task interference \cite{chen2025gradient_survey, sener2018multi, Dsidri2012MultiplegradientDA_mgda, yu2020gradient_pcgrad, liu2021conflict_cagrad}. GradNorm \cite{chen2018gradnorm} balances magnitudes of task gradients by modulating them based on task training rates. However, it disregards gradient directional conflicts and relies on hyperparameter tuning. PCGrad \cite{yu2020gradient_pcgrad} modifies task gradients by conditionally projecting a task's gradient onto the orthogonal sub-space of another task’s gradient only when a conflict is detected based on negative cosine similarity, and does not consider orderings for the objectives. In our work, we project energy minimization gradients onto the orthogonal sub-space of reward maximization gradients at all times thereby prioritizing reward maximization over energy minimization. We show empirically that directly applying PCGrad-style conditional projection leads to sub-optimal policies in Sec.~\ref{sec: simulation_expts}. Grad-Similarity \cite{du2018adapting} uses cosine similarity between task gradient vectors to estimate task-relatedness. However, the objective of Grad-Similarity is to select auxiliary tasks that can optimize a single main task objective whereas our work attempts to jointly optimize multiple task objectives. CAGrad \cite{liu2021conflict_cagrad} uses a hyperparameter to find a vector that maximizes worst-case improvement while staying within a ball around the average gradient of tasks. Our method is hyperparameter free and does not explicitly constrain the resulting vector to be proximate to the average gradient.  GradDrop \cite{chen2020just_graddrop} randomly drops conflicting gradient components based on their signs to reduce interference. However, in contrast to our work, GradDrop does not consider gradient magnitudes and task importance. Xu et al. \cite{Xu2023CompositeML} perform adaptive gradient updates to ensure that no single objective dominates, whereas, our work assumes a priority ordering where gradients of the task reward are more important than that of energy.

\textbf{Energy optimization in Legged Robots}: Nai et al.~\cite{nai2025fine} train a surrogate model to predict energy consumption by deploying a pre-trained policy and collecting real-world power usage data. This surrogate model is then used to define an additional energy-based reward, which guides the fine-tuning of the original policy. The authors adopt an iterative refinement loop involving real-world data collection, surrogate model training, and policy fine-tuning. In contrast, our approach focuses on directly optimizing for energy efficiency in simulation, followed by transfer to the real world—eliminating the need for a learned energy model as well as an iterative refinement process.
Fu et al. \cite{fu2021minimizing} train reinforcement learning policies to minimize energy—formulated as mechanical work—alongside task and survival rewards, and observe that the resulting behaviors resemble natural gaits in biological animals. While they manually tune the reward weights, we propose to automatically learn a policy that minimizes energy while preserving task performance. 
Mahankali et al.~\cite{mahankali2024maximizing} formulate the energy-performance trade-off as a constrained optimization problem by maintaining two separate policies: one for the combined task and energy objective ($\pi$) and another for the task objective alone ($\pi'$). While $\pi'$ is optimized conventionally, the updates to $\pi$ are penalized whenever its task reward falls below that of $\pi'$, with the penalty scaled by a learnable parameter $\alpha$. This parameter $\alpha$ is dynamically adjusted to keep the returns of $\pi$ and $\pi'$ close. Although their problem setting aligns with ours, we take a different approach: instead of maintaining two policies, we introduce two critics and optimize a single policy.

\section{Methodology}
\vspace{-5pt}
We formulate our problem as a Multi-Objective Markov Decision Process (MOMDP) \cite{hayes2022practical} defined by the tuple $(S, A, T, R, \gamma, \mu_0)$ where $S$ and $A$ are the state and action spaces, $T(s'|s, a)$ is the transition dynamics, $R(s)$ is a vector-valued reward function representing scalar rewards for each objective, $\gamma$ is the discount factor, and $\mu_0(s)$ is the start state distribution. In our setting, we consider the reward function to be 2-dimensional -- returning both a task reward $r(s)$ and energy consumption $e(s)$ at state $s$. Our overall goal will be to minimize energy consumption without sacrificing task reward.

\subsection{Preliminaries}
\vspace{-5pt}
\label{sec: prelim}

While our proposed method is applicable to any policy-gradient algorithm, we present results on two popular actor-critic \cite{konda1999actor} algorithms -- Soft Actor Critic (SAC) \cite{haarnoja2018soft} and Proximal Policy Optimization (PPO) \cite{schulman2017ppo} -- and focus our discussion here on SAC to describe our approach. For a matching discussion of PPO, see the Appendix \ref{apdx: mo_ppo}.

\textbf{Soft Actor Critic (SAC).} In standard single-objective settings, SAC aims to maximize both expected returns and policy entropy. Following the actor-critic framework, SAC learns a  scalar action-value function $Q_\phi(s, a)$ and a stochastic policy $\pi_\theta(a|s)$. The critic network $Q_\phi(s,a)$ is trained to minimize 
\begin{equation}
\mathcal{L}_Q = \left( Q_\phi(s, a) - \left( r + \gamma Q_{\bar{\phi}}(s', a') - \alpha \log \pi_\theta(a' \mid s') \right) \right)^2
\label{eq: qloss}
\end{equation}
where $a' \sim \pi_\theta(.|s')$, $\alpha$ is an entropy coefficient, and $Q_{\bar{\phi}}$ is the target Q-network  implemented as an exponential moving average of $Q_{\phi}$. The policy function $\pi_\theta(a|s)$ is trained to minimize 
\begin{equation}
\mathcal{L}_\pi = \mathbb{E}_{a \sim \pi_\theta} \left[ \alpha \log \pi_\theta(a \mid s) - Q_\phi(s, a) \right].
\end{equation}
Both the critic and policy networks are trained simultaneously.

\textbf{Multi-Objective SAC.} For our dual objective setting, we employ two critic networks -- one for estimating the task action-value function $Q_\phi^r(s, a)$ and the other for the energy action-value function $Q_{\phi_e}^{e}(s, a)$. Each critic can be trained independently following Eq.~\ref{eq: qloss} for rewards and energy consumption. These critics can share parameters; however, we implement them independently here.

For deriving a policy in multi-objective settings, a common approach is to introduce (or learn) a utility function $U$ that maps the vector-valued Q-function to a scalar by encoding some notion of the relative importance of each objective. This reduces the policy-learning problem back to that of a single-objective MDP where the resulting utility can be maximized. 
A common choice of utility function when it comes to energy minimization is a simple linear combination parameterized by a trade-off parameter $\lambda$. For SAC, this corresponds to the following actor objective:
\begin{equation}
\label{eqn:combined_actor}
    \begin{aligned}
        \mathcal{L}_\pi = \mathbb{E}_{a \sim \pi_\theta} \left[ \alpha \log \pi_\theta(a \mid s) - Q^r_\phi(s, a)+ \lambda Q_{\phi_e}^{e}(s, a)\right]
    \end{aligned}
\end{equation}
Larger values of $\lambda$ will more strongly move the policy's distribution away from high energy states and actions. However, it is often difficult to set the trade-off parameter $\lambda$ \emph{a priori} for a new task. Too high and the policy may fail to learn or produce sluggish behaviors. Too low and energy may not be effectively optimized. Furthermore,  optimizing combinations of conflicting objectives can result in sub-optimal learning \cite{yu2020gradient_pcgrad} -- which certainly include task reward and energy expenditure for highly dynamic robot control tasks like locomotion.

\subsection{Projecting Energy Gradients (\ours{})} 
\vspace{-5pt}
\label{sec: ours_methodology}
To alleviate this issue, we introduce our method \ourcolor{\textbf{P}}rojecting \ourcolor{\textbf{E}}nergy \ourcolor{\textbf{Grad}}ients (\ours{}). To motivate our proposed method, we consider a multi-objective setting in which energy minimization is a subordinate objective that should be optimized \emph{only} when doing so does not interfere with reward maximization. To start, let us separate the actor objective into two losses corresponding to the task reward and energy reduction, denoting these as
\begin{eqnarray}
\mathcal{L}_R(\theta) {=} \mathbb{E}_{a \sim \pi_\theta} \left[ \alpha \log \pi_\theta(a \mid s) - Q^r_\phi(s, a) \right]~~~\mbox{and}~~~ 
\mathcal{L}_E(\theta) {=} \mathbb{E}_{a \sim \pi_\theta} \left[ \alpha \log \pi_\theta(a \mid s) + Q_{\phi_e}^{e}(s, a) \right]
\end{eqnarray}
For a small change $d$ in policy parameters $\theta$, the reward objective for the resulting policy under a first-order Taylor approximation can be written as
\begin{eqnarray}
\mathcal{L}_R(\theta + d) \approx \mathcal{L}_R(\theta) + g_R^Td
\end{eqnarray}
where $g_R$ is the gradient of the reward loss with respect to network parameters evaluated at $\theta$, i.e., $g_R = \nabla_\theta \mathcal{L}_R(\theta)$. From this, standard gradient descent algorithms setting $d=-\alpha g_R$ can be derived where $\alpha$ is a learning rate hyperparameter. More to our point however, this approximation also implies that a small change in a direction orthogonal to $g_R$ results in \emph{no} change to the reward loss function -- corresponding to movement along a level-set of the approximated reward loss hyperplane. Naturally, this is only valid in a small region around $\theta$ where the approximation holds.

This suggests a straight-forward algorithm in which the energy consumption loss is minimized only by shifting parameters in this orthogonal space of the reward loss gradient. By the same reasoning as above, changing $\theta$ by adding $-g_E=-\nabla_\theta \mathcal{L}_E(\theta)$ would move to minimize energy; however, $-g_E$ may have components that would also modify $\mathcal{L}_R$. Instead, we consider the orthogonal projection of $g_E$ onto $g_R$ denoted as $g_{E_{\perp R}}$, taking the overall descent direction $d$ as
\begin{eqnarray}
d=-\alpha g_R - \beta~g_{E_{\perp R}} = -\alpha g_R - \beta\left(g_E - \frac{g_R^Tg_E}{g_R^Tg_R} g_R\right)
\end{eqnarray}
Rather than tuning $\beta$ as an independent learning rate hyperparameter, we define it adaptively as
\begin{eqnarray}
\beta = \alpha*\min\left(1, \frac{\mid\mid g_R\mid\mid_2}{\mid\mid g_{E_{\perp R}}\mid\mid_2} \right)
\end{eqnarray}
such that the norm of $\beta g_{E_{\perp R}}$ is no greater than the norm of $\alpha g_R$. We provide an empirical justification for this choice, finding it to work well across multiple settings.%

\textbf{Implementing \ours{}.} Alg.~\ref{alg: algorithm} outlines a practical implementation of our proposed approach. After performing backward passes for each policy loss component, energy loss gradients are directly adjusted via projection and conditionally rescaled. Rather than setting $\alpha$ as a learning rate directly, we pass the updated gradient direction $g_R + g_{E_{\perp R}}$ to any choice of optimizer. This is applied at each step of training. Applying this algorithm represents a relatively small code change in existing RL frameworks but does incur the cost of a second backward pass to compute $g_R$ and $g_E$ separately.

\begin{algorithm}[h]
\caption{\ours{}}
\begin{algorithmic}[1]

\Require Policy ($\pi$) parameters $\theta$, task minibatch $\mathcal{B} = \{ \mathcal{T}_k \}$
\State $g_R \gets \nabla_\theta\mathcal{L}_R(\theta)$\Comment{Compute task reward loss gradient from batch}\vspace{2.5pt}

\State $g_E \gets \nabla_\theta\mathcal{L}_E(\theta)$\Comment{Compute energy expenditure loss gradient from batch}\vspace{2.5pt}

\State $g_{E_{\perp R}} \gets g_E - \frac{g_R^Tg_E}{g_R^Tg_R} g_R$ \Comment{Compute orthogonal projection}\vspace{2.5pt}

\If{$\left\lVert g_{E_{\perp R}}\right\rVert_2  ~~>~~\left\lVert g_R\right\rVert_2$}\vspace{2.5pt}
    \State $g_{E_{\perp R}} = g_{E_{\perp R}}\frac{\left\lVert g_R\right\rVert_2}{\left\lVert g_{E_{\perp R}}\right\rVert_2}$ \Comment{Rescale if larger norm than reward gradient}\vspace{2.5pt}
\EndIf\vspace{2.5pt}
\State \textbf{return} update $\Delta \theta = g_R + g_{E_{\perp R}}$ \Comment{Pass returned value as gradient to optimizer}
\end{algorithmic}
\label{alg: algorithm}
\end{algorithm}
\vspace{-15pt}

\section{Experiments}
\vspace{-5pt}
We center our experiments and discussion around the following questions: \textbf{(Q1)} Can \ours{} minimize energy while retaining task performance across various environments and tasks? (Sec. \ref{sec: simulation_expts}); and \textbf{(Q2)} Can we use \ours{} to reduce battery usage on a real robot? (Sec. \ref{sec: sim2real})

\textbf{Formulation of energy.} Prior works have used various formulations of energy such as torque penalty and mechanical work. In our experiments, we choose the sum of absolute torques applied to actuated motors as a proxy for energy. Concretely, we compute the energy function as %
$e(s) = \sum_{m=1}^M |\tau_m|$
where $M$ is the total number of actuated motors of the robot and $\tau_m$ is the torque applied to the $m$th motor. This formulation aims to minimize battery current draw rather than system energy. In simulation, current $I$ is unmeasurable but proportional to motor torque $\tau$, and taking $|\tau|$ captures draw regardless of direction. On contrary, mechanical power $(\tau.\omega)$ depends on the velocity, and can underestimate $I$ during high-torque, low-motion phases. Thus, $\sum |\tau|$ better reflects battery load. However, we note that this ignores differences in motors and gear ratios that impact current draw. If these are heterogeneous and known for an embodiment, appropriate coefficients may be included.\looseness=-1

\subsection{Simulation Experiments}
\vspace{-5pt}
\label{sec: simulation_expts}
\textbf{Setting.} For our simulation experiments, we use two environments based on the MuJoCo \cite{mujoco} physics simulator -- DM-Control suite \cite{tunyasuvunakool2020dm_control} and  HumanoidBench \cite{sferrazza2024humanoidbench}.

\begin{figure}[t]
\centering
\includegraphics[width=0.95\linewidth]{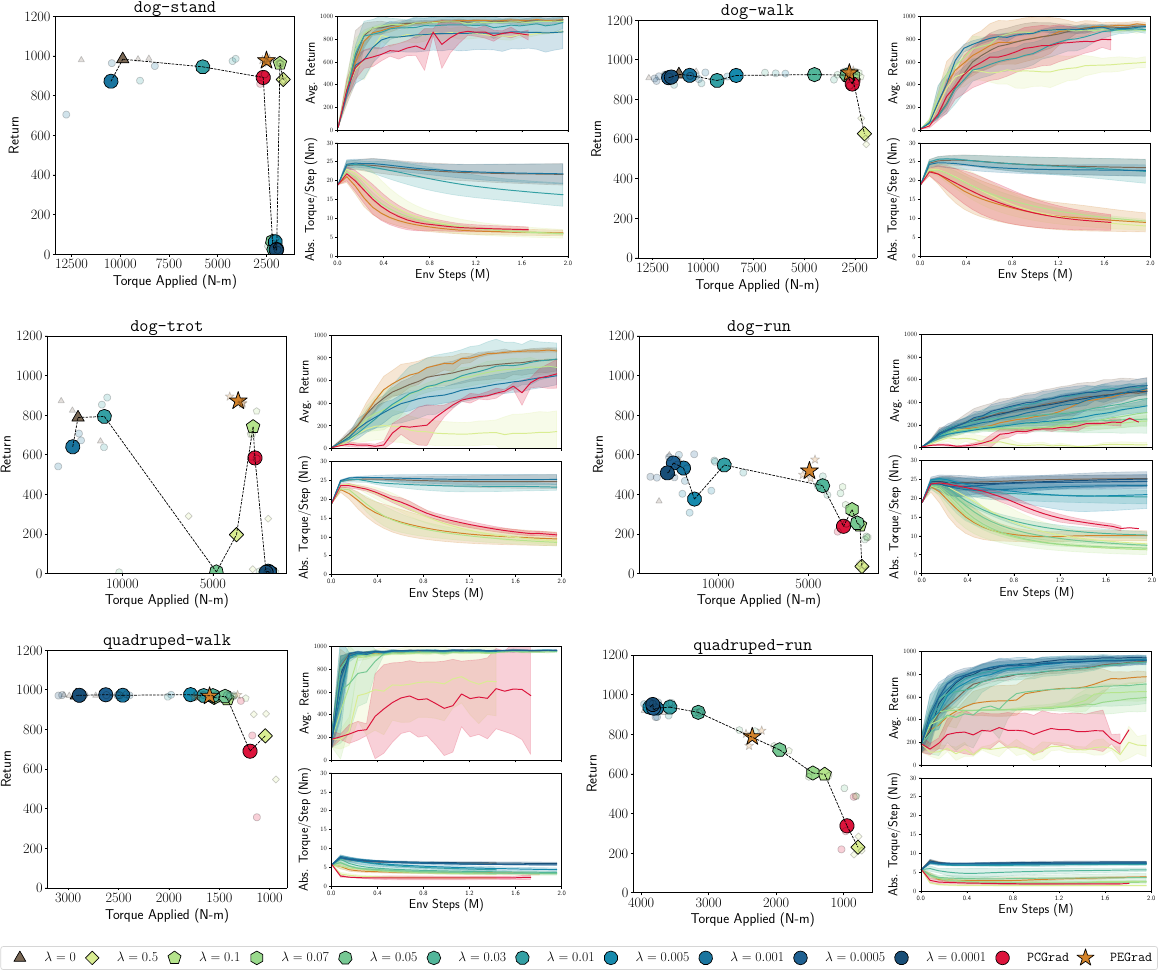}
  \caption{\texttt{DMControl Suite} Results: We show results on six tasks from \texttt{DMControl} suite \cite{tunyasuvunakool2020dm_control}. Low applied torque and higher returns are better.Across all tasks, \ours{} achieves high task performant policies that are also energy-efficient. For 4 out of 6 tasks, \ours{} achieves results beyond the Pareto front identified by adjusting $\lambda$. }
  \label{fig:01_dmcontrol}
  \vspace{-5pt}
\end{figure}

\texttt{DM-Control}: We consider two robot embodiments \texttt{Quadruped} and \texttt{Dog} and conduct our experiments on \texttt{Quadruped-\{Walk, Run\}}, and \texttt{Dog-\{Stand, Walk, Run, Trot\}}. Tasks on the \texttt{Dog} embodiment are the most challenging in the suite owing to their 228-dim state space and 38 dimensional action space. Policy actions correspond directly to motor torques.

\texttt{HumanoidBench}: The \texttt{HumanoidBench} benchmark consists of 27 whole body control tasks. We focus on four fundamental locomotion tasks \texttt{\{Stand, Walk, Run, Sit\}}, all of them trained on Unitree H1 humanoid robot in simulation. %
All the tasks have state dimension of 51 and action dimension 19. Policy actions correspond directly to motor torques.

 For the experiments on \texttt{DM-Control} suite and \texttt{HumanoidBench} we use SAC as the base RL algorithm that is implemented using LeanRL \cite{leanrl_github} (a PyTorch library based on CleanRL \cite{huang2022cleanrl}). A list of hyperparameters and network architectures for critics and actor is described in Appendix \ref{apdx: SAC}.
 
\textbf{Baselines.} To validate our method, we compare \ours{} against the following baselines:
\begin{compactenum}[\hspace{0.1cm}-]
\item\texttt{Base}: The base SAC implementation has a single critic and is trained on default environment reward functions. For \texttt{DM-Control} this corresponds to an energy unconstrained policy (denoted as $\lambda{=}0$), whereas \texttt{HumanoidBench} includes some default energy penalties (denoted as \texttt{Base}).\\[-8pt] %
\item\texttt{Multi-Objective ($\lambda{=}X$)}: This baseline includes two critics as described in Sec.~\ref{sec: prelim} with a factor $\lambda$ in the policy loss to trade off between task reward and energy efficiency as in Eq.~\ref{eqn:combined_actor}. We use energy as defined above for all environments and remove default energy penalties from \texttt{HumanoidBench} rewards. We select $\lambda \in \{0.001, 0.01, 0.1, 0.5\}$ for our experiments.\\[-8pt]
\item\texttt{PCGrad$^+$} \cite{yu2020gradient_pcgrad}: While originally proposed for multi-objective problems where all objectives are equally important, we adapt PCGrad to our setting where reward takes precedence over energy. The resulting method will first subtract the parallel component of $g_E$ when $g_E$ and $g_R$ are conflicting i.e $g_R ^\top g_E < 0$, but otherwise aggregates both gradients directly. We build on the public PyTorch implementation of this work \cite{Pytorch-PCGrad}.
\end{compactenum}

\textbf{Results.} For each task setting, we provide three result plots -- (i) a \textit{Pareto front} showing our energy formulation vs.~return for converged policies; (ii) a \textit{sample efficiency} plot showing environment steps vs.~return; and (iii) an \textit{energy efficiency} plot showing environment steps vs.~our energy formulation.  All results shown are run for 1.5-2M steps and run for 3 seeds. Shaded areas in (ii)/(iii) are 95\% bootstrapped CIs. For (i), mean results are plotted with individual seeds as shaded markers.

We report results for the \texttt{DMControl} suite of tasks in Fig.~\ref{fig:01_dmcontrol}.  We observe that \ours{} consistently produces energy efficient policies -- achieving high rewards on-par with the unconstrained \texttt{Base} ($\lambda=0$) policy while executing with significantly reduced torques. Interestingly, we observe that on both the \texttt{quadruped} and the more energetic \texttt{dog-run} and \texttt{dog-trot} tasks, \texttt{PCGrad$^+$} leads to sub-optimal policies -- sacrificing substantial returns for gains in energy efficiency. %

\begin{figure}[t]
\centering
\includegraphics[width=0.90\linewidth]{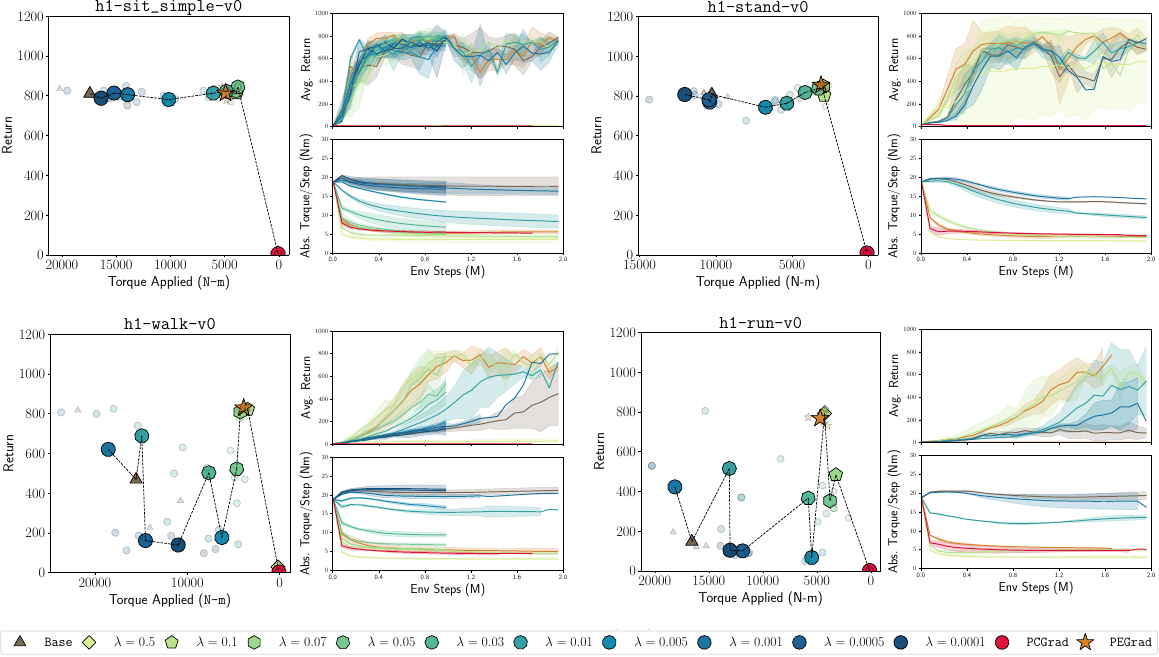}
  \caption{\texttt{HumanoidBench} Results: We show results on four tasks from \texttt{HumanoidBench} suite \cite{sferrazza2024humanoidbench}. Low applied torque and higher returns are better. Across all tasks, \ours{} achieves highly performant policies that are also energy-efficient with \ours{}. Further, energy minimization also improves sample-efficiency on \texttt{h1-run-v0} and \texttt{h1-walk-v0} tasks.}
  \label{fig:02_humanoidbench}
  \vspace{-11pt}
\end{figure}

We report results for the \texttt{HumanoidBench} suite of tasks in Fig.~\ref{fig:02_humanoidbench}. We observe \ours{} not only leads to lower-energy and highly-performant policies but also shows \textit{significant} sample-efficiency gains compared to the \texttt{Base} on \texttt{run} and \texttt{walk} tasks. The $\lambda=0.1$ models also exhibit this phenomenon -- indicating that minimizing energy objectives can lead to improved sample-efficiency in some RL tasks. Interestingly, we find that \texttt{PCGrad$^+$} over-optimizes energy to the point of achieving nearly no return for any of the four tasks. The corresponding policy effectively falls without much actuation. Speculatively, we attribute this to \texttt{PCGrad$^+$}'s lack of any adaptive gradient scaling analogous to $\beta$.%

\subsection{Sim2Real on Go2 Quadruped}
\vspace{-5pt}
\label{sec: sim2real}

\textbf{Experimental Setting.} For our Sim2Real experiments, we train a standing and walking (SaW) controller for the Unitree Go2 quadruped in simulation using IsaacLab \cite{mittal2023orbit} and deploy it in the real world. We train using PPO \cite{rudin2022learning} on  flat terrain with standard domain randomization, and adopt the Adversarial Motion Prior (AMP) framework \cite{Peng_2021} for "natural-looking" gaits which has been adopted extensively in robotic settings \cite{wang2023learning, huang2025think, tang2024humanmimic, escontrela2022adversarial, zargarbashi2024robotkeyframing, li2023learning}. AMP models a motion prior over a set of demonstration trajectories by training a discriminator  that learns to distinguish between real trajectories and the robot's generated motions. The robot's control policy is trained in an adversarial fashion to this discriminator, thereby encouraging the robot to produce motions that are similar to the collected trajectories. To build a dataset of demonstrations, we use the factory-issued Go2 controller to collect a 90 second sequence of standing and walking trajectories. The trajectories consist of 24 dimensional states (joint positions and velocities) and 12-dimensional actions (joint-delta positions with respect to the canonical pose of the robot) recorded at 50Hz. During training, the robot’s policy receives two rewards: one for tracking desired velocities (task performance), and another \textit{style reward} from the AMP discriminator, encouraging gaits similar to the factory-issued controller. We provide more details about the specific architecture and details of AMP in Section \ref{apdx: AMP}. 

For the \texttt{standing} task, we command 0 velocity and compute metrics over 20 seconds. For the \texttt{walking} task, we command the robot with a velocity of $0.5$m/s in the forward direction to cover a distance of 12 feet ($\sim 3.66$m). For both, we report the current drawn as well as the net absolute torque applied in that time. For \texttt{standing}, we allow policies an initial burnin period during which we do not collect metrics. For \texttt{walking}, we likewise %
ignore the initial and final period where the commanded speed of the robot changes due to start-up and slow-down. These filtering steps are to account for input delays and the effects of switching from default to learned controllers. %

\textbf{Baselines.} We compare the energy efficiency of \ours{} against the factory-issued Go2 controller (denoted as \texttt{Factory}) as well as  \texttt{AMP+PPO $\lambda$} baselines with $\lambda=0$ and $\lambda=0.0002$ which corresponds to the best baseline policy (on task reward) that we could achieve with tuning this hyperparameter manually. We note that task performance across all three policies is similar in simulation.

\begin{table}[t]
\begin{center}
\setlength{\tabcolsep}{1.2em}
\begin{tabular}{c|c|c|c|c}
        \toprule
        SaW Controller & \multicolumn{2}{c}{\shortstack{\texttt{Standing}}} & \multicolumn{2}{c}{\shortstack{\texttt{Walking}}}\\[3pt]
          {} & \shortstack{Current \\ Drawn \tiny{\texttt{(mA)}} } & \shortstack{Net Torque \\[-3pt] Applied \tiny{\texttt{(Nm)}}} & \shortstack{Current \\ Drawn \tiny{\texttt{(mA)}}} & \shortstack{Net Torque \\[-3pt] Applied \tiny{\texttt{(Nm)}}}\\
        \midrule \midrule
           Factory & 4.029 \PM 0.005 & 3.97 \PM 0.001  & 6.46 \PM 0.19 & 4.58 \PM 0.04 \\
           \midrule
           \texttt{AMP+PPO $\lambda=0$} & 3.473 \PM 0.005 & 3.47 \PM 0.083 & \textcolor{red}{\ding{55}} & \textcolor{red}{\ding{55}} \\
           \texttt{AMP+PPO $\lambda=0.0002$} & \textbf{2.389 \PM 0.170}  & \textbf{2.52 \PM 0.006} & 7.04 \PM 0.90 & 4.68 \PM 0.01 \\
        AMP+PPO \ours{} & \textbf{2.533 \PM 0.022} & \textbf{2.45 \PM 0.105} & \textbf{5.65 \PM 0.45} & \textbf{3.94 \PM 0.01}\\\bottomrule
\end{tabular}
\end{center}
\caption{Current and Torque usage in the real-world: We compare Unitree's \texttt{Factory} controller and \texttt{AMP+PPO $\lambda$} baselines against \ours{} for \texttt{Standing} and \texttt{Walking} tasks and report current drawn and net torque applied. We find that \ours{} is ${\sim}20\%$ more efficient than the tuned multi-objective \texttt{AMP+PPO $\lambda{=}0.0002$} on the task of \texttt{walking} and has a comparable performance on \texttt{standing}.}
\label{tab: sim2real}
\vspace{-11pt}
\end{table}

\textbf{Results.} Across both \texttt{standing} and \texttt{walking} tasks shown in Tab.~\ref{tab: sim2real}, we find policies trained with \ours{} lead to significantly lower currents being drawn from the battery as well as reduced torques being applied compared to the factory policy. When deployed for the task of \texttt{walking}, the \texttt{AMP+PPO $\lambda=0$} policy was unsuccessful in completing the 12 feet distance and showcased unsafe behaviors such as taking large jumps -- because of which we decided to omit it from the evaluation (marked with \textcolor{red}{\ding{55}} in the table). Interestingly, we  find that the tuned torque penalty (\texttt{AMP+PPO $\lambda=0.0002$}) and \ours{} achieve similar performance on all metrics for the \texttt{standing} task; however,  we observe a \textit{significant} gap in performance between the two on the \texttt{walking} task. \ours{} is $19.74\%$ more efficient on current usage and $15.8\%$ on net torques applied. We would like to point out that these numbers denote \textit{per-timestep} current and torque usage and can show significant difference especially when a robot operates for a longer period of time.

\textbf{Time Analysis.} On a stand-alone RTX 4090 GPU, both \ours{} and the baseline \texttt{AMP+PPO} take ${\sim}$4.5 hours to complete 10k iterations, achieving $\sim$62k steps per second (SPS). On an HPC cluster with an L40 GPU, \ours{} takes 8.4 hours ($\sim$35k SPS), while the baseline takes 7.6 hours ($\sim$37k SPS). This difference likely stems from other bottlenecks such as simulation overhead rather than policy update, particularly given that most RL control policies use relatively shallow networks.

\section{Conclusion}
\vspace{-5pt}
\label{ref:conclusion}
In this work, we introduced \ours{}, a method for incorporating energy minimization into RL-based robot control without compromising task performance. By projecting energy gradients orthogonal to task reward gradients, \ours{} avoids the need for sensitive hyperparameter tuning between the two objectives and prioritizes task success over energy objectives. Our extensive evaluations across \texttt{DMControl}, \texttt{HumanoidBench}, and real-world deployments on the Unitree Go2 quadruped demonstrate that \ours{} consistently achieves significant reductions in energy usage—up to $64\%$ in simulation—while retaining or improving policy performance. Moreover, \ours{} enables Sim2Real transfer of energy-efficient policies, offering practical gains in battery usage and robot longevity.\looseness=-1

\section{Limitations}
\vspace{-5pt}
\label{sec:discussion}
\textbf{Interplay Between \ours{} and Style Rewards.} Throughout this project, we found it challenging to combine the energy minimization objective with \textit{style rewards}. In tasks like \texttt{DMControl} and \texttt{HumanoidBench}, which lack explicit style constraints, \ours{} effectively learned energy-efficient and task-performant policies. However, when we initially conducted experiments without AMP-based style rewards i.e traditional reward engineering with style-enforcing rewards such as \textit{feet-air time} \cite{van2024revisiting}, \textit{mirror loss} \cite{mittal2024symmetry}, and other penalties not directly tied to energy—e.g., \textit{joint deviation}, \textit{base acceleration}, and \textit{action rate}—we observed interesting behaviors that were sometimes infeasible or unsafe for Sim2Real transfer. For instance, in the absence of a constraint on base height, one policy learned to maximize its base height as a way to reduce per-step torque, which resulted in a constrained set of allowable joint displacements. This, in turn, led to shorter, rapid steps so as to maintain the task reward. Another policy that was trained without any style reward robot learned a dragging behavior where 2 legs are actively moving and the rest are dragging themselves, minimizing the overall torque. Additional cases are discussed in Appendix~\ref{apdx:sim2real}. These instances depict the challenges while training with multiple reward functions and one potential way to alleviate this could be to recursively use \ours{} in the order of priority of objectives, with each of the objectives having a separate critic. We leave this as a future direction of our work.

\section{Acknowledgements}
\vspace{-5pt}
SP would like to thank Mohit Gadde and Aayam Shrestha for their help with the real-world experiments, Hunter Brown for his inputs on energy formulation, Rob Yelle for his support of GPU resources during the deadline, and ViRL lab members for their feedback on the draft of this work.  This work is supported in part by NSF Award No. 2321851. The views and conclusions contained herein are those of the authors and should not be interpreted as necessarily representing the official policies or endorsements, either expressed or implied, of the U.S. Government, or any sponsor.

\bibliography{references}  %

\newpage 
\section{Appendix}
\subsection{Multi-Objective PPO}
\label{apdx: mo_ppo}
\textbf{Proximal Policy Optimization (PPO).} PPO \cite{schulman2017ppo} is a widely used on-policy algorithm that optimizes a surrogate objective based on the clipped probability ratio between the new and old policies. Specifically, PPO seeks to maximize the expected advantage while limiting the deviation from the previous policy using a clipped objective:
\begin{equation}
\mathcal{L}_{\text{PPO}} = \mathbb{E}_t \left[ \min\left( r_t(\theta) \hat{A}_t, \text{clip}(r_t(\theta), 1 - \epsilon, 1 + \epsilon)\hat{A}_t \right) \right]
\end{equation}
where $r_t(\theta) = \frac{\pi_\theta(a_t|s_t)}{\pi_{\theta_{\text{old}}}(a_t|s_t)}$ is the probability ratio between the current and old policies, $\epsilon$ is a hyperparameter controlling the clip range, and $\hat{A}_t$ is an estimate of the advantage function. The value function is learned via regression to the discounted returns, and the policy and value networks are trained simultaneously. An entropy bonus is often added to the objective to encourage exploration.

\textbf{Multi-Objective PPO.} In our multi-objective setting, we extend PPO to optimize over both task performance and energy consumption. We maintain two separate reward signals: one for task reward $r^r$ and another for energy usage $r^e$, and compute separate advantage estimates $\hat{A}^r_t$ and $\hat{A}^e_t$ accordingly. To combine them, we use a linear scalarization approach with a trade-off parameter $\lambda$, forming a scalarized advantage:
\begin{equation}
\hat{A}^{\text{total}}_t = \hat{A}^r_t - \lambda \hat{A}^e_t
\end{equation}
This scalarized advantage is used in the PPO surrogate objective:
\begin{equation}
\mathcal{L}_{\text{MO-PPO}} = \mathbb{E}_t \left[ \min\left( r_t(\theta) \hat{A}^{\text{total}}_t, \text{clip}(r_t(\theta), 1 - \epsilon, 1 + \epsilon)\hat{A}^{\text{total}}_t \right) \right]
\end{equation}
As with SAC, setting the trade-off parameter $\lambda$ is crucial: larger values more strongly penalize energy usage but can degrade task performance or convergence.

\subsection{Environments}
\label{apdx: environment}
\begin{figure}[h]
\centering
\includegraphics[width=0.5\textwidth]{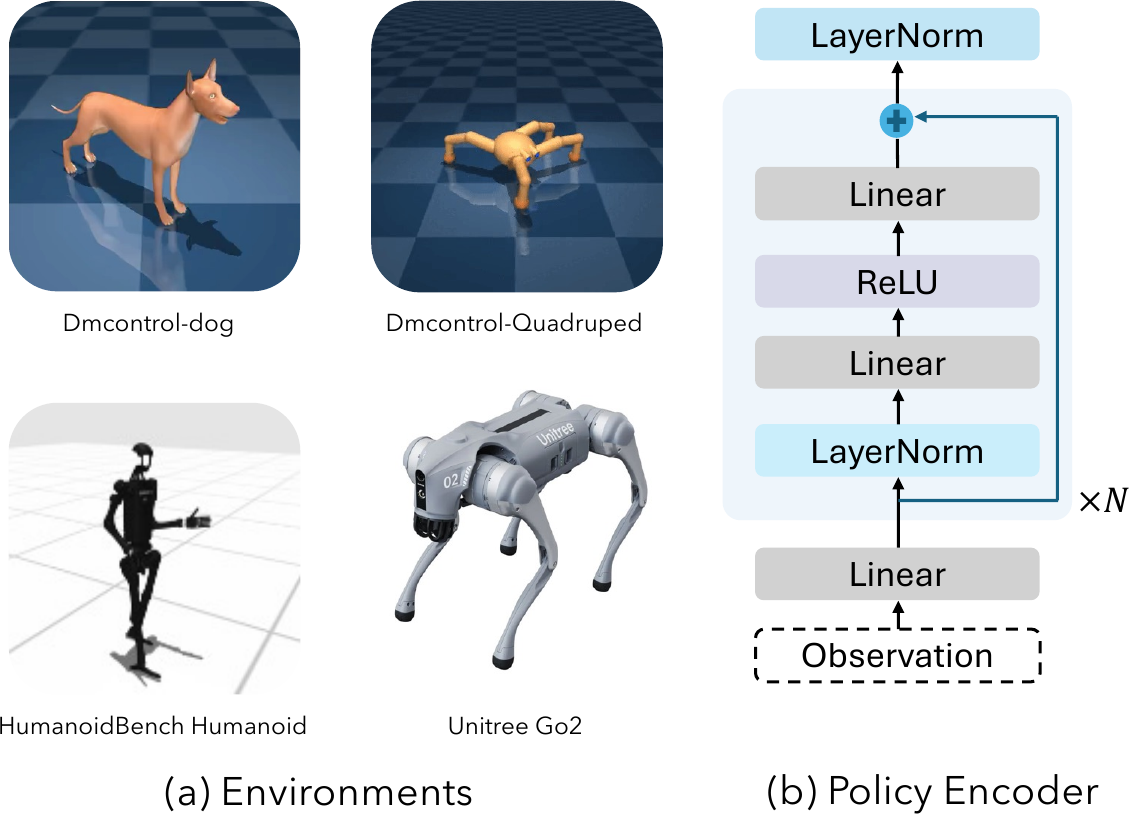}
  \caption{\textit{a) Environments} -- We consider 4 locomotion environments -- (i) \texttt{DMControl-Dog}, (ii) \texttt{DMControl-Quadruped}, (iii) \texttt{HumanoidBench-H1 Humanoid}, and (iv) \texttt{Unitree Go2}. We test \ours{} on 10 simulated tasks with \texttt{DMControl} and \texttt{HumanoidBench} environments and conduct a Sim2Real evaluation of standing and walking tasks with Unitree Go2 quadruped. \textit{b) Policy Encoder} -- We choose a SimBa-like \cite{lee2025simba} architecture that has shown sample-efficiency and better task performance by carefully selecting the architecture encoder for Actor and Critic in Deep-RL algorithms.}
  \label{fig:appdx-tasks}
  \vspace{-11pt}
\end{figure}

We consider 10 simulation tasks (six from \texttt{DMControl} suite \citep{tunyasuvunakool2020dm_control} and 4 locomotion tasks from \texttt{HumanoidBench} benchmark \cite{sferrazza2024humanoidbench} as shown in Figure \ref{fig:appdx-tasks}. The dimensionality of the observation space and action space is mentioned in Table \ref{tab:appdx-state-action-space}.

\begin{table}[h!]
\centering
\parbox{\textwidth}{
\centering
\begin{tabular}{lccc}
\toprule
\textbf{Task} & \textbf{Observation dim} & \textbf{Action dim} \\ \midrule
\texttt{quadruped-walk} & $78$ & $12$ \\
\texttt{quadruped-run} & $78$ & $12$ \\
\texttt{dog-stand} & $223$ & $38$ \\ 
\texttt{dog-walk}& $223$ & $38$ \\ 
\texttt{dog-trot} & $223$ & $38$ \\ 
\texttt{dog-run} & $223$ & $38$ \\ 
\midrule
\texttt{h1-stand} & $51$ & $19$ \\ 
\texttt{h1-sit\_simple} & $51$  & $19$ \\ 
\texttt{h1-walk} & $51$ & $19$ \\ 
\texttt{h1-run} & $51$ & $19$ \\ 
\midrule
\texttt{Go2-Sim2Real} & $48$ & $12$ \\ \bottomrule
\end{tabular}%
\caption{State and Action spaces of each of the tasks.}
\label{tab:appdx-state-action-space}
}
\end{table}

\subsection{AMP + PPO Training}
\label{apdx: AMP}

\textbf{Dataset Collection.} Demonstration trajectories are collected using a factory-provided controller by manually operating the robot via a joystick for the locomotion task, capturing joint positions and joint velocities at 50 Hz. This yields a 24-dimensional observation vector (joint positions and velocities) corresponding to 12 actuated joints. The trajectory spans 20 seconds, and we further augment the dataset by extracting all possible sliding windows of 0.2 seconds to get a large batch of trajectories. 

The discriminator is implemented as a fully connected multilayer perception (MLP) with two hidden layers of 1024 and 256 units, respectively, with ELU activations. It receives the 24-dimensional proprioception as input and outputs a single logit that denotes the confidence that the original observation is from real demonstrations. The discriminator is trained jointly with the policy, with updates occurring at every learning step.

The style reward is derived by applying a sigmoid activation to the discriminator's logits to obtain the probability that a sample is classified as real. Denoting this probability by \( p = \sigma(\text{logits}) \), the style reward is then computed as the negative log-probability of the discriminator predicting ``real'':

\begin{align}
        r_{style} &= -\log\left( \max\left(1 - \sigma(\text{logits}), \epsilon\right) \right),
\end{align}

where $\epsilon$ is a small constant to prevent numerical instability. This reward encourages the policy to provide behaviors the indistinguishable from real demonstration. The style reward is scaled by a factor $\lambda_{style}$ and combined with the task reward and penalties.

The task reward promotes velocity tracking locomotion by minimizing the L2 deviation between commanded and actual linear (X, Y) and angular velocities. Additionally, for the Baseline (\texttt{AMP+PPO}), we incorporate auxiliary penalty terms such as linear motion in the Z-axis and angular velocities about the X and Y axes (roll and pitch rates) to promote stable locomotion. Termination conditions, such as minimum base height and base contact, are also added. We describe the exact reward structure used to train GO2 SaW (Standing and Walking) controller with Adverserial Motion Priors (AMP) \cite{Peng_2021} in Table \ref{tab:appdx-amp-reward}.

\begin{table}[t]
\centering
\caption{Individual task, style and penalty rewards for \texttt{AMP+PPO} baseline.}
\parbox{\textwidth}{
\centering
\begin{tabular}{lccc}
\toprule
\textbf{Reward Term} & \textbf{Definition} & \textbf{Weighting} \\ \midrule
Base linear $x,y$ velocity & $e^{-\frac{\| \mathbf{v}_{cmd} - \mathbf{v}_{current} \|^2}{0.25}}$ & 1.5 \\
Base angular yaw velocity & $e^{-\frac{\|w_{cmd} - w_{current}\|^2}{0.25}}$ & 0.75 \\
AMP Style reward & $-\log\left( \max\left(1 - \sigma(\text{logits}), \epsilon\right) \right)$ & 0.4 \\
Feet air time & 
$\begin{cases}
\sum\limits_{k \in \mathcal{F}} (t_{\text{air},k} - t_{\text{thresh}}) \cdot \mathbf{1}_{\text{fc},k}
& \text{if } \|\mathbf{v}_{\text{cmd}}\| > \delta \\
0 & \text{otherwise}
\end{cases}$ 
& 0.25 \\

\hline
Vertical velocity penalty & $\|v_z\|^2$ & -2.0 \\
Roll/pitch velocity penalty & $\|w_{xy}\|^2$ & -0.05 \\
Orientation penalty & $\|\theta_{\text{rp}}\|^2$ & -2.5 \\
Torque penalty (not used in \ours{}) & $\|\tau\|^2$ & -0.0002 \\
\bottomrule
\end{tabular}%

\vspace{0.5ex}
\noindent\textbf{Notes.}
\begin{itemize}
    \item $\mathbf{v}_{\text{cmd}}$ is the commanded linear velocity.
    \item $\mathbf{v}_{\text{current}}$ is the base linear velocity.
    \item $\mathbf{w}_{\text{cmd}}$ is the commanded angular velocity.
    \item $\mathbf{w}_{\text{cmd}}$ is the base yaw angular velocity.
    \item $\mathbf{logits}$ is a discriminator network output.
    \item $\mathbf{\epsilon}$ is a numerical stability term to avoid computing $\log$ of zero.
    \item $\sigma$ is the sigmoid function that computes the probability from the discriminator $\mathbf{logits}$.
    \item $\mathcal{F}$ denotes the set of feet. In this case, it denotes four legs: front-left, front-right, rear-left, and rear-right.
    \item $t_{\text{air},k}$ is the air-time duration for foot $k$.
    \item $t_{\text{thresh}}$ is the threshold air-time for the penalty.
    \item $\mathbf{1}_{\text{fc},k}$ is 1 if foot $k$ just made contact with the ground, 0 otherwise.
    \item $\theta_{\text{rp}}$ is the roll and pitch angles of the robot base.
    \item $\delta$ is the minimum movement threshold to enable feet-air-time reward. This is required to disable the feet-air-time reward when the command is near zero.
    \item $\tau$ is the applied torque for each of the joints
\end{itemize}

}

\label{tab:appdx-amp-reward}

\end{table}

\subsection{SAC Hyperparameters and Architecture}
\label{apdx: SAC}

We show our encoder architecture for policy as well as for the critic in Figure \ref{apdx: environment} where we base our model on SimBa \cite{lee2025simba}. In this section we list out the hyperparameters for our SAC algorithm used on \texttt{DMControl} and \texttt{HumanoidBench} training in Table \ref{tbl:sac_hyperparams}.

\begin{table}[ht]
\centering
\caption{\textbf{Hyperparameters Table.} The hyperparameters listed below are used consistently
across all baseline and lagrange versions of SAC, unless stated otherwise.}
\small
\vspace{4mm}
\label{tbl:sac_hyperparams}
\resizebox{0.83\textwidth}{!}{

\begin{tabular}{llll}
\toprule
& \textbf{Hyperparameter} & \textbf{Notation} & \textbf{Value} \\
\midrule
\multirow{4}{*}{\textbf{Common}}
& Discount factor & $\gamma$ & $0.99$ \\
& Replay buffer capacity     & - & $1$M \\
& Buffer sampling            & - & Uniform \\
& Batch size                 & - & $256$ \\
\midrule
\multirow{4}{*}{\textbf{Actor}} 
& Number of SimBa-like blocks & - & 1 \\
& Hidden dimension & - & $\{256/512\}$ (\texttt{DMControl} / \texttt{HumanoidBench}) \\
& Entropy coefficient & $\alpha$ & $0.2$ \\
& Update frequency & - & $2$\\ \midrule
\multirow{3}{*}{\textbf{Critic}} 
& Number of SimBa-like blocks & - & 2 \\
& Hidden dimension & - & $\{256/512\}$ (\texttt{DMControl} / \texttt{HumanoidBench} \\
& Update frequency & - & $1$\\ 

\midrule
\multirow{5}{*}{\textbf{Optimizer}} 
& Optimizer          & - & Adam \\
& Optimizer momentum & $(\beta_1, \beta_2)$ & (0.9, 0.999) \\
& Weight Decay       & - & 0.0 \\
& Policy LR      & - & 3e-4 \\
& Critic LR       & - & 1e-3 \\
\bottomrule
\end{tabular}
}
\end{table} %

\subsection{Energy Formulation}
\label{apdx:energy_formulation}
In addition to formulating energy as sum of absolute joint torques, we consider mechanical power $(\tau . \omega)$. It is important to note that mechanical power can underestimate the current $I$ drawn from the battery during high-torque, low-motion phases.

\begin{figure}[!h]
\centering
\includegraphics[width=0.5\textwidth]{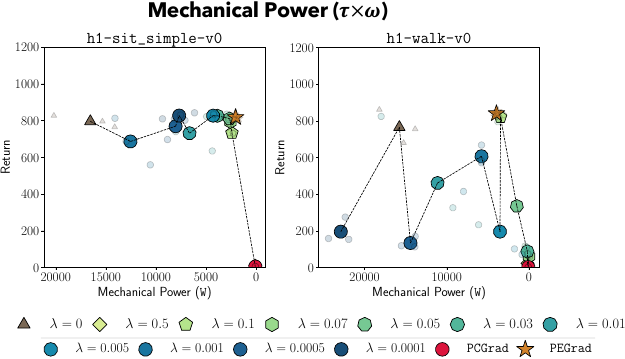}
  \caption{We show results on HumanoidBench's \texttt{sit\_simple} and \texttt{walk} tasks with mechanical power $(\tau . \omega)$ as the energy formulation. We observe that \ours{} continues to achieve energy-efficient and task-performant policies even with a different formulation.}
  \label{fig:appdx-energy-formulation}
  \vspace{-11pt}
\end{figure}

\subsection{Sim2Real Discussion}
\label{apdx:sim2real}

In this section, we first show some additional quantitative results that we conducted on Unitree Go2 robot for Standing and Walking (SaW) tasks. Next we highlight some additional examples of challenges we encountered during the process of performing Sim2Real experiments with style rewards.

\begin{figure}[h]
\centering
\includegraphics[width=0.8\textwidth]{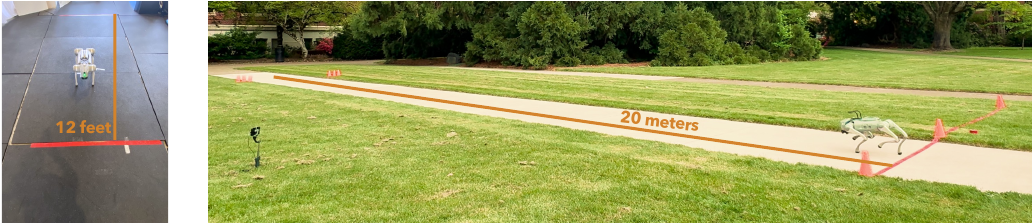}
  \caption{\textbf{Real world setup.} \textit{(left)} Go2 in Lab-setting: We consider a rubber mat flooring  terrain on which the dataset for AMP has been collected. For the results in Table \ref{tab: sim2real}, we consider a distance of 12 feet ($\sim 3.66$ meters) (marked in \textcolor{red}{red}) where the robot is commanded a velocity of 0.5 m/s. \textit{Right} Outdoor setting where we test our policies on the \texttt{concrete} pathway as well as the adjacent \texttt{grass} terrain. For this, we consider a larger distance of 20 meters (distance between the cones).}
  \label{fig:appdx-real-world-setup}
\end{figure}

\textbf{Additional Sim2Real results.}

We deployed the trained SaW policies on different terrains to see if the lower energy consumption that was trained using reference trajectories that were collected in a lab-setting in Section \ref{sec: sim2real}, transferred similarly to outdoor terrains such as grass and concrete. All the trained Go2 policies were trained with dynamics randomization on a \textit{flat terrain}. It is important to note that these set of experiments are \textit{not} to evaluate the generalization capability of the policy to various terrains -- rather to see if there is any significant deviation in terms of energy consumption of \ours{} versus the baselines. We report our results in Table \ref{tab: apdx_sim2real}.

Across both the scenarios, we find \ours{} to perform significantly better in terms of both current drawn as well as the net torque applied. Specifically, we find \ours{} to be better than the finetuned \texttt{AMP+PPO $\lambda=0.0002$} baseline by $\sim 24.5\%$ on concrete and $\sim 7.68\%$ on grass. We suspect this reduction of energy consumption even for the baseline on grass terrain is because grass can absorb part of the impact, reducing the need for active damping or stabilization from the motors. On the other hand, the average current drawn on \texttt{Concrete} is similar to the lab-setting where the robot was deployed on a rubber sheet. One important observation was that in grassy \& concrete terrains the policy sometimes ends up at a lower speed than commanded. We attribute this sim2real gap to the fact that all our GO2 policies were trained on flat terrain. We hypothesize that further training of policies on a diverse terrain would alleviate this issue.

\begin{table}[t]
\begin{center}
\setlength{\tabcolsep}{1.2em}
\begin{tabular}{c|c|c|c|c}
        \toprule
        SaW Controller & \multicolumn{2}{c}{\shortstack{\texttt{Walking (Concrete)}}} & \multicolumn{2}{c}{\shortstack{\texttt{Walking (Grass)}}}\\[3pt]
          {} & \shortstack{Current \\ Drawn \tiny{\texttt{(mA)}} } & \shortstack{Net Torque \\[-3pt] Applied \tiny{\texttt{(Nm)}}} & \shortstack{Current \\ Drawn \tiny{\texttt{(mA)}}} & \shortstack{Net Torque \\[-3pt] Applied \tiny{\texttt{(Nm)}}}\\
        \midrule \midrule
           Factory & 5.430 \PM  0.007 & 3.97 \PM 0.009  & 4.875 \PM 0.181 & 3.72 \PM 0.050  \\
           \midrule
           \texttt{AMP+PPO $\lambda=0.0002$} & 6.905 \PM 0.753  & 4.47 \PM 0.115 & 4.669 \PM 0.384 & 3.71 \PM 0.077  \\
        AMP+PPO \ours{} & \textbf{5.208 \PM 0.399} & \textbf{3.45 \PM 0.014} & \textbf{4.310 \PM 0.002} & \textbf{3.26 \PM 0.005} \\\bottomrule
\end{tabular}
\end{center}
\caption{Current and Torque usage on different terrains: We compare Unitree's \texttt{Factory} controller and \texttt{AMP+PPO $\lambda$} baselines against \ours{} for \texttt{Walking} task on \texttt{Concrete} and \texttt{Grass} and report current drawn and net torque applied. We find that \ours{} is ${\sim}24.5\%$ more efficient than the tuned multi-objective \texttt{AMP+PPO $\lambda{=}0.0002$} on \texttt{Concrete} terrain and ${\sim}7.68\%$ on \texttt{Grass} terrain. Results averaged over two trials.}
\label{tab: apdx_sim2real}
\end{table}

\textbf{Additional examples for interplay between \ours{} and Style Rewards}
 
As discussed in our limitations (Section \ref{sec:discussion}), one of the challenges we faced with minimizing energy was to learn a behavior alongside style rewards. In addition to the examples mentioned in the Limitations section, we often observed hopping or trotting behaviors in simulation when working with different style reward structures. With a reward structure the adds a mirror loss \cite{van2024revisiting} which is common in several humanoid and quadruped locomotion literature to encourage the gaits to be symmetric about the sagittal plane, the robot learns a hopping behavior (we show a video of this in the supplementary video).

\end{document}